\pdfoutput=1

\documentclass[11pt,a4paper]{article}
\usepackage[colorlinks,linkcolor=blue]{hyperref}
\usepackage[hyperref]{acl2020}
\usepackage{times}
\usepackage{latexsym}

\usepackage{microtype}

\usepackage{CJKutf8}
\usepackage{amsmath}
\usepackage{graphicx}
\usepackage{float}
\usepackage{subfigure}
\usepackage{booktabs}
\usepackage{amssymb}
\usepackage{threeparttable}
\usepackage[disable]{todonotes}
\usepackage{pgfplots}
\usepackage{scalefnt}

\aclfinalcopy % Uncomment this line for the final submission
\title{FLAT: Chinese NER Using Flat-Lattice Transformer}

\author{
Xiaonan Li, Hang Yan, Xipeng Qiu\thanks{\ \  Corresponding author.}, Xuanjing Huang \\
Shanghai Key Laboratory of Intelligent Information Processing, Fudan University \\
School of Computer Science, Fudan University \\
lixiaonan\_xdu@outlook.com,\quad \{hyan19, xpqiu, xjhuang\}@fudan.edu.cn
}

\date{}

\begin{document}
\maketitle

\begin{CJK}{UTF8}{gbsn}
  % \maketitle

  \begin{abstract}
  % unperfectly ready

%  Chinese named entity recognition (NER) usually adopts character-level sequence labeling models to avoid the error prorogation of word segmentation.
% Recently, the character-word lattice structure has been proved to be effective for Chinese named entity recognition (NER) by incorporating the word information. However, since the lattice structure is complex and dynamic, the lattice-based models are hard to fully utilize the parallel computation of GPUs and usually have a low inference speed. In this paper, we propose \textbf{FLAT}: \textbf{F}lat-\textbf{LA}ttice \textbf{T}ransformer for Chinese NER, which converts the lattice structure into a flat structure consisting of spans. Each span corresponds to a character or latent word and its position in the original lattice. With the power of Transformer and well-designed position encoding, FLAT can fully leverage the lattice information and has an excellent parallel ability. Experiments on four datasets show FLAT outperforms other lexicon-based models in performance and efficiency.

Recently, the character-word lattice structure has been proved to be effective for Chinese named entity recognition (NER) by incorporating the word information. However, since the lattice structure is complex and dynamic, most existing lattice-based models are hard to fully utilize the parallel computation of GPUs and usually have a low inference-speed. In this paper, we propose \textbf{FLAT}: \textbf{F}lat-\textbf{LA}ttice \textbf{T}ransformer for Chinese NER, which converts the lattice structure into a flat structure consisting of spans. Each span corresponds to a character or latent word and its position in the original lattice. With the power of Transformer and well-designed position encoding, FLAT can fully leverage the lattice information and has an excellent parallelization ability. Experiments on four datasets show FLAT outperforms other lexicon-based models in performance and efficiency.
  \end{abstract}

  \section{Introduction}
  %第一段，NER是什么，有什么用，中文NER的困难之处
  % Named entity recognition aims to identify predefined semantic spans, including their location and classification.

  % Named entity recognition (NER) seeks to locate the entities' spans and classify them into predefined semantic categories, such as person names, organizations, locations, etc.
  Named entity recognition (NER) plays an indispensable role in many downstream natural language processing (NLP) tasks \cite{chen-etal-2015-event,Diefenbach2018}.
  %, such as relation extraction \cite{Bunescu:2005:SPD:1220575.1220666}, event extraction \cite{chen-etal-2015-event}, question answering system \cite{Diefenbach2018}.
  Compared with English NER \cite{lample-etal-2016-neural,DBLP:journals/corr/YangZD17aa,DBLP:journals/corr/abs-1709-04109,sun2020sparsing}, Chinese NER is more difficult since it usually involves word segmentation. %In detail, named entity boundaries are always word boundaries. But inherently Chinese has much less boundary information than English.\par One intuitive way of alleviating the boundary problem is converting Chinese NER into the pipeline of word segmentation and word-level sequence labeling \citep{10.1007/978-3-319-75477-2_9,He:2017:UMC:3298023.3298036}. However the segmentation → NER pipeline cannot avoid error propagation problem. It is fatal to Chinese NER \cite{zhang-etal-2006-word} Because wrong segmented entity boundaries lead to NER error, while named entity is probably OOV which is a significant source of segmentation error. The pipeline method is also sensitive to the segmentation criteria. However, only using character information will suffer from disregard of word information, including boundary and semantic representation. %The former is really important for chinese NER as explained before. For the latter, in Chinese a word's meaning could be irrelevant to the characters which constitute it. For instance, the word ``沙发'', the transliteration of sofa, is composed of characters ``沙(sand)'' and ``发(send)''. Since they are irrelevant, it's theoretically difficult to get accurate representation of such words.\par
  \begin{figure}[!t] %H为当前位置，!htb为忽略美学标准，htbp为浮动图形
    \centering %图片居中
      \subfigure[Lattice.]{
        \includegraphics[width=0.47\textwidth]{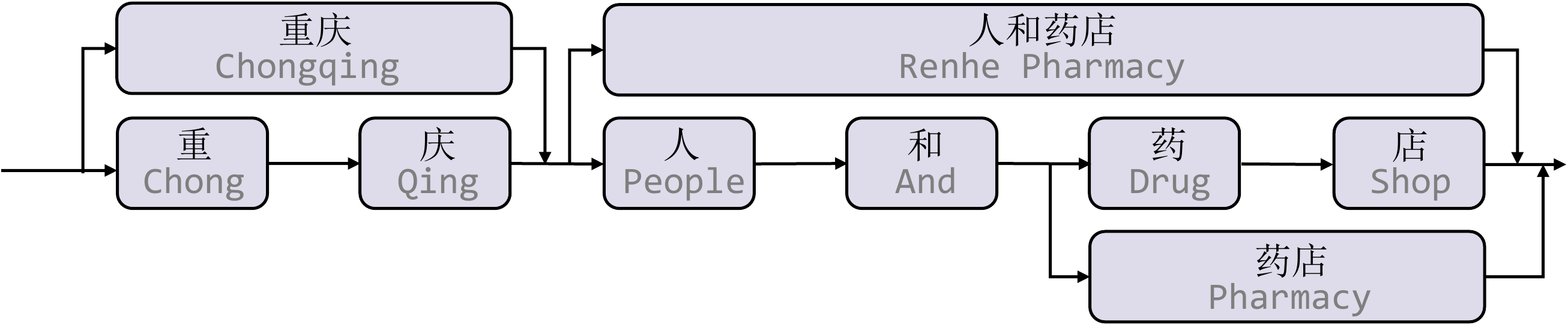}
        \label{lattice}
      }

      \subfigure[Lattice LSTM.]{
        \includegraphics[width=0.47\textwidth]{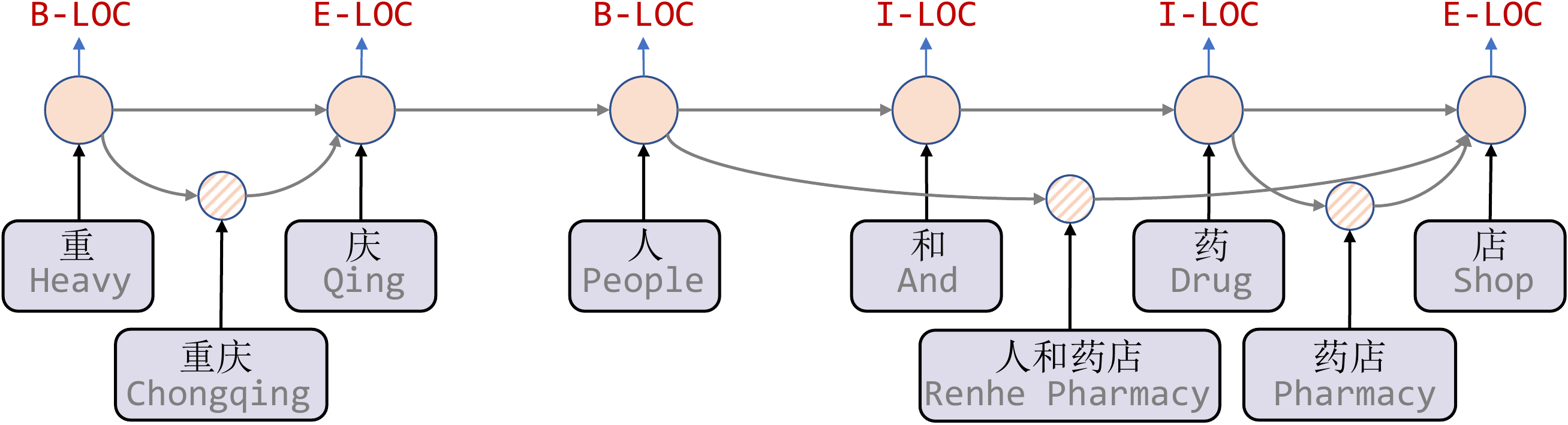}
        \label{lattice lstm}
      }
      \subfigure[Flat-Lattice Transformer.]{
        \includegraphics[width=0.47\textwidth]{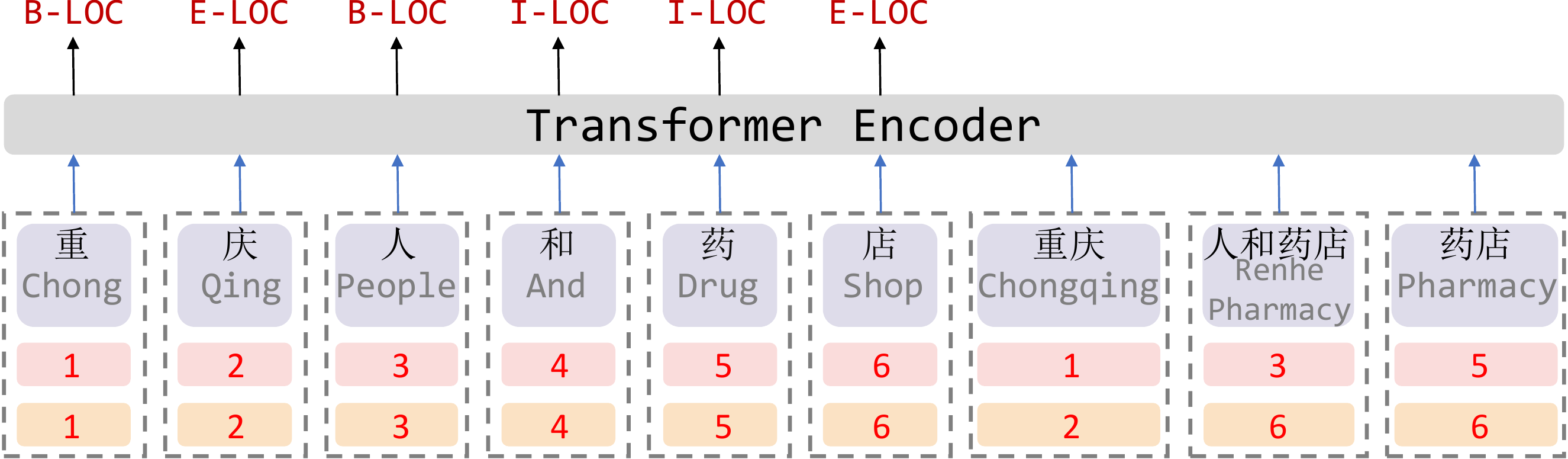}
        \label{flat lattice}
      }
    % \begin{subfigure}
    % \includegraphics[width=0.5\textwidth]{img/zhaohai_clip.pdf}%插入图片，[]中设置图片大小，{}中是图片文件名
    % \end{subfigure}
    % \begin{subfigure}
    % \includegraphics[width=0.5\textwidth]{img/us_clip.pdf}%插入图片，[]中设置图片大小，{}中是图片文件名
    % \end{subfigure}
    \caption{ While lattice LSTM indicates lattice structure by dynamically adjusting its structure, FLAT only needs to leverage the span position encoding. In \ref{flat lattice},\ \includegraphics[width=.25cm]{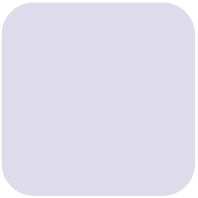}, \includegraphics[width=.25cm]{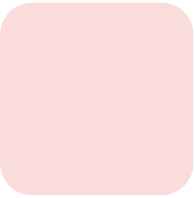}, \includegraphics[width=.25cm]{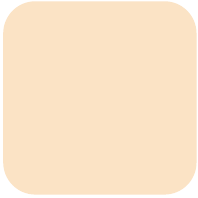} denotes tokens, heads and tails, respectively.
    % \todo[inline]{show lstm and transformer,refer BERT fig}
    } %最终文档中希望显示的图片标题
    \label{figure1}
  \end{figure}
  \par Recently, the lattice structure has been proved to have a great benefit to utilize the word information and avoid the error propagation of word segmentation~\cite{DBLP:journals/corr/abs-1805-02023}. We can match a sentence with a lexicon to obtain the latent words in it, and then we get a lattice like in Figure \ref{lattice}. The lattice is a directed acyclic graph, where each node is a character or a latent word. The lattice includes a sequence of characters and potential words in the sentence. They are not ordered sequentially, and the word's first character and last character determine its position.
  % Some word like ``人和药店(Renhe Pharmacy)'' can be important for figuring out the geographic entity ``重庆(Chongqing)'' instead of the organization entity ``重庆人(Chongqing People)''.
  Some words in lattice may be important for NER. For example, in Figure \ref{lattice}, ``人和药店(Renhe Pharmacy)'' can be used to distinguish between the geographic entity ``重庆(Chongqing)'' and the organization entity ``重庆人(Chongqing People)''.

  There are two lines of methods to leverage the lattice.
  (1) One line is to design a model to be compatible with lattice input, such as lattice LSTM \cite{DBLP:journals/corr/abs-1805-02023} and LR-CNN \cite{Gui:2019:CCN:3367722.3367738}. In lattice LSTM, an extra word cell is employed to encode the potential words, and attention mechanism is used to fuse variable-number nodes at each position, as in Figure \ref{lattice lstm}.
  %For a potential word, there is a skip-edge from its first character to its last character, and this is how lattice LSTM indicates the structure of inputs. %For example, ''人和药店`` is connected from position '3` to '6`.
  LR-CNN uses CNN to encode potential words at different window sizes. However, RNN and CNN are hard to model long-distance dependencies \cite{NIPS2017_7181}, which may be useful in NER, such as coreference \cite{stanislawek-etal-2019-named}. Due to the dynamic lattice structure, these methods cannot fully utilize the parallel computation of GPU.
  (2) Another line is to convert lattice into graph and use a graph neural network (GNN) to encode it, such as Lexicon-based Graph Network (LGN) \cite{gui-etal-2019-lexicon} and Collaborative Graph Network (CGN) \cite{sui-etal-2019-leverage}. While sequential structure is still important for NER and graph is general counterpart, their gap is not negligible. These methods need to use LSTM as the bottom encoder to carry the sequential inductive bias, which makes the model complicated.

  In this paper, we propose \textbf{FLAT}: \textbf{F}lat \textbf{LA}ttice \textbf{T}ransformer for Chinese NER.
  Transformer \cite{NIPS2017_7181} adopts fully-connected self-attention to model the long-distance dependencies in a sequence. To keep the position information, Transformer introduces the position representation for each token in the sequence. Inspired by the idea of position representation, we design an ingenious position encoding for the lattice-structure, as shown in Figure \ref{flat lattice}.
  In detail, we assign two positional indices for a token (character or word): head position and tail position, by which we can reconstruct a lattice from a set of tokens.
  %For example in Figure \ref{flat lattice}, the head position and tail position of token ``人和药店(Renhe Pharmacy)'' are 3 and 6. The head position and tail position is the same for the character token.
  Thus, we can directly use Transformer to fully model the lattice input. The self-attention mechanism of Transformer enables characters to directly interact with any potential word, including self-matched words. To a character, its self-matched words denote words which include it. For example, in Figure \ref{lattice}, self-matched words of ``药 (Drug)'' are ``人和药店(Renhe Pharmacy)'' and ``药店 (Pharmacy)''\cite{sui-etal-2019-leverage}.
  Experimental results show our model outperforms other lexicon-based methods on the performance and inference-speed. Our code will be released at \href{https://github.com/LeeSureman/Flat-Lattice-Transformer}{https://github.com/LeeSureman/Flat-Lattice-Transformer}.
  \todo[inline]{两层的效果比一层的差，所以就不能说我们利用了词与词之间的交互了，在一层时字并不能够利用词与词互相交互的信息}
  \section{Background}
  In this section, we briefly introduce the Transformer architecture. Focusing on the NER task, we only discuss the Transformer encoder. It is composed of self-attention and feedforward network (FFN) layers. Each sublayer is followed by residual connection and layer normalization.
  FFN is a position-wise multi-layer Perceptron with non-linear transformation. Transformer performs self-attention over the sequence by $H$ heads of attention individually and then concatenates the result of $H$ heads. For simplicity, we ignore the head index in the following formula. The result of per head is calculated as:
  % \begin{equation}
  %   \operatorname{Att}(\mathbf{A}, \mathbf{V})=\operatorname{softmax}
  %   % \left(\frac{\mathbf{Q_i} \mathbf{K_i}^{\mathrm{T_i}}}{\sqrt{\mathrm{d_{head}}}}\right)
  % (\mathbf{A})
  %   \mathbf{V},\\
  %   \end{equation}
  % \begin{equation}
  %   \mathbf{A_{ij}} = \left(\frac{\mathbf{Q_{i}} \mathbf{K_{j}}^{\mathrm{T}}}{\sqrt{\mathrm{d_{head}}}}\right)
  % \end{equation}
  % \begin{equation}
  %   [\mathbf{Q},\mathbf{K},\mathbf{V}] = X [\mathbf{W}^{Q},\mathbf{W}^{K},\mathbf{W}^V]
  % \end{equation}

\vspace{-1em}
  {\small
  \begin{align}
    \label{weighted_sum_att}
    \operatorname{Att}(\mathbf{A}, \mathbf{V})&=\operatorname{softmax}
    % \left(\frac{\mathbf{Q_i} \mathbf{K_i}^{\mathrm{T_i}}}{\sqrt{\mathrm{d_{head}}}}\right)
  (\mathbf{A})
    \mathbf{V},\\
    \mathbf{A_{ij}} &= \left(\frac{\mathbf{Q_{i}} \mathbf{K_{j}}^{\mathrm{T}}}{\sqrt{\mathrm{d_{head}}}}\right), \\
    [\mathbf{Q},\mathbf{K},\mathbf{V}] &= E_{x} [\mathbf{W}_{q},\mathbf{W}_{k},\mathbf{W}_{v}],
  \end{align}}%
  %   \begin{equation}
  %   \mathbf{A_{ij}} = \left(\frac{\mathbf{Q_i}} \mathbf{K_j}^{\mathrm{T}}{\sqrt{\mathrm{d_{head}}}}\right)
  % \end{equation}
  % \begin{equation}
  %   \operatorname{Att_{(i)}}(\mathbf{A_{(i)}}, \mathbf{V_{(i)}})=\operatorname{softmax}
  %   % \left(\frac{\mathbf{Q_i} \mathbf{K_i}^{\mathrm{T_i}}}{\sqrt{\mathrm{d_{head}}}}\right)
  % (\mathbf{A_{(i)}})
  %   \mathbf{V_{(i)}},\\
  %   \end{equation}
  % SinceIn each head, the result of attention is calculated as:
  % \begin{equation}
  %   \begin{align}
  %     (a + b)^3  &= (a + b) (a + b)^2        \\
  %                &= (a + b)(a^2 + 2ab + b^2) \\
  %                &= a^3 + 3a^2b + 3ab^2 + b^3
  %   \end{align}
  % \end{equation}
  % \begin{equation}
  %   \operatorname{Att_{(i)}}(\mathbf{A_{(i)}}, \mathbf{V_{(i)}})=\operatorname{softmax}
  %   % \left(\frac{\mathbf{Q_i} \mathbf{K_i}^{\mathrm{T_i}}}{\sqrt{\mathrm{d_{head}}}}\right)
  % (\mathbf{A_{(i)}})
  %   \mathbf{V_{(i)}},\\
  %   \end{equation}
  % \begin{equation}
  %   \mathbf{A_{(i)}} = \left(\frac{\mathbf{Q_{(i)}} \mathbf{K_{(i)}}^{\mathrm{T}}}{\sqrt{\mathrm{d_{head}}}}\right)
  % \end{equation}
  % \begin{equation}
  %   [\mathbf{Q_{(i)}},\mathbf{K_{(i)}},\mathbf{V_{(i)}}] = X [\mathbf{W_{(i)}}^{Q},\mathbf{W_{(i)}}^{K},\mathbf{W_{(i)}}^V]
  % \end{equation}
  where $E$ is the token embedding lookup table or the output of last Transformer layer. $\mathbf{W}_\mathrm{q},\mathbf{W}_\mathrm{k},\mathbf{W}_\mathrm{v} \in \mathbb{R}^{d_{model} \times d_{head}}$ are learnable parameters, and $d_{model} = H \times d_{head}$, $d_{head}$ is the dimension of each head.

  %The denominator $\sqrt{d_{head}}$ is used for scaling the attention score over the whole sequence to prevent extremely small gradients.
  %But \citet{Yan2019TENERAT} found this trick is not suitable for NER, so we drop the scale factor.
  The vanilla Transformer also uses absolute position encoding to capture the sequential information. Inspired by \citet{Yan2019TENERAT}, we think commutativity of the vector inner dot will cause the loss of directionality in self-attention. Therefore, we consider the relative position of lattice also significant for NER.
  \begin{figure}[tbp] %H为当前位置，!htb为忽略美学标准，htbp为浮动图形
    \centering %图片居中
      \includegraphics[width=0.47\textwidth]{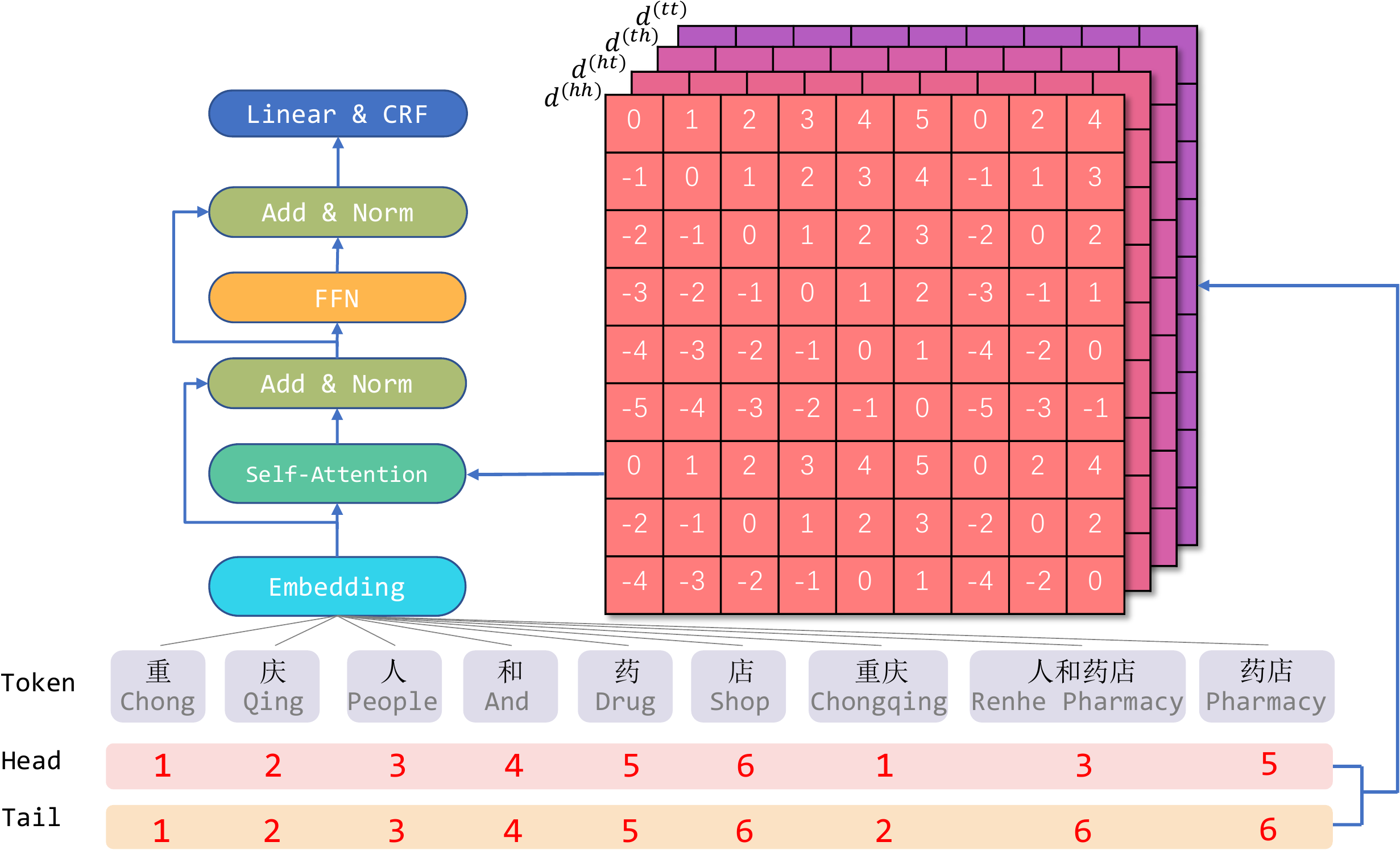}
      % \label{lattice-graph}
      % \subfigure[lattice LSTM]{
      %   \includegraphics[width=0.43\textwidth]{img/model-crop.pdf}
      %   \label{lattice-graph}
      % }
    % \begin{subfigure}
    % \includegraphics[width=0.5\textwidth]{img/zhaohai_clip.pdf}%插入图片，[]中设置图片大小，{}中是图片文件名
    % \end{subfigure}
    % \begin{subfigure}
    % \includegraphics[width=0.5\textwidth]{img/us_clip.pdf}%插入图片，[]中设置图片大小，{}中是图片文件名
    % \end{subfigure}
    \caption{The overall architecture of FLAT.} %最终文档中希望显示的图片标题
    \label{model}
  \end{figure}
  \todo[inline]{上面这段关于相对的，可简略，放到model里去}

  \section{Model}
  \subsection{Converting Lattice into Flat Structure}
  % After getting a lattice from a character with a lexicon, we can flatten it into flat counterpart. Formally, we transform the lattice into tokens ($x_[1:m]$), heads ($h[1:m]$), tails ($t[1:m]$), where m is the number of nodes in the lattice and the index of nodes does not represent their sequential order. The tokens are the nodes' content such as `重(heavy)', `庆(qing)', `重庆(Chongqing)', etc. The heads and tails are the index of nodes' first and last characters. For example, the head and tail of `人和药店(Renhe Pharmacy)' is the index of `人(people)' and `店(shop)', 3 and 6. For one character, its head and tail are same. And we take the flat lattice: tokens, heads and tails as the input of our model. The flat lattice can be regarded as a set of spans and one span corresponds to a token, a head and a tail.
  After getting a lattice from characters with a lexicon, we can flatten it into flat counterpart.
  %We regard characters and words in the lattice as tokens, and their positions are determined by their first and last characters. Formally, the flat-lattice can be defined as a set of spans, and a span corresponds to a token, a head and a tail. The token is a character or word, such as ``重(Chong)'', ``庆(Qing)'', ``重庆(Chongqing)'', etc. The head and tail denote the position index of the token's first and last characters, and they indicate the position of the token in the original lattice.
The flat-lattice can be defined as a set of spans, and a span corresponds to a token, a head and a tail, like in Figure \ref{flat lattice}. The token is a character or word. The head and tail denote the position index of the token's first and last characters in the original sequence, and they indicate the position of the token in the lattice. For the character, its head and tail are the same.
  % For example, the head and tail of token ``人和药店(Renhe Pharmacy)'' in figure \ref{lattice} are 3 and 6. To a character, its head and tail is the same such as the head and tail of ``庆(Qing)'' are both 2. We take tokens' embedding as the input of our model, use head and tail to indicate their original structure.
  There is a simple algorithm to recover flat-lattice into its original structure. We can first take the token which has the same head and tail, to construct the character sequence. Then we use other tokens (words) with their heads and tails to build skip-paths. Since our transformation is recoverable, we assume flat-lattice can maintain the original structure of lattice.
  \todo[inline]{需不需要说明分别对char和word进行不同的线性变换 1.让他们变为同一维度 2.防止他们预训练的表示空间不同}

  % \subsection{Heterogeneous Embedding}
  % We use different pretrained embedding resource for character and words respectively while we take them . To prevent the gap of two pretrained spaces damaging the effect of pretraining, we use two different linear transformation to make model can quickly align these two pretrained spaces as follows:
  % \begin{equation}\small
  %   E(x)=\left\{
  %   \begin{aligned}
  %     \mathbf{W}_{c\_proj\ }&E^{(c)} (x)\\
  %     \mathbf{W}_{w\_proj}&E^{(w)} (x)
  %   \end{aligned}
  %   \right.,
  %   \end{equation}
  % % where \begin{small}$E^{(c)}$\end{small}, \begin{small}$E^{(w)}$\end{small} are character and word embedding lookup table, and \begin{small}$\mathbf{W}_{c\_proj}$\end{small}, \begin{small}$\mathbf{W}_{w\_proj}$\end{small} are learnable parameters.

  % where $E^{(c)}$, $E^{(w)}$ are character and word embedding lookup table, and $\mathbf{W}_{c\_proj}$, $\mathbf{W}_{w\_proj}$ are learnable parameters.

  \subsection{Relative Position Encoding of Spans}
  % The flat-lattice structure consists of spans with different lengths. To encode the interactions among spans, we propose the relative position encoding of spans. For two spans $x_i$ and $x_j$ in the lattice, there are three kinds of relations between them: intersection, inclusion and separation. We use a simple mechanism to capture the complex relations. It can not only represent the relation between two tokens, but also indicate more detailed information, such as how many characters are overlapped. Let $head[i]$ and $tail[i]$ denote the head and tail position of span $x_i$. Four kinds of relative distances can be used to indicate the relation between $x_i$ and $x_j$. They can be calculated as:
  The flat-lattice structure consists of spans with different lengths. To encode the interactions among spans, we propose the relative position encoding of spans. For two spans $x_i$ and $x_j$ in the lattice, there are three kinds of relations between them: intersection, inclusion and separation, determined by their heads and tails. Instead of directly encoding these three kinds of relations, we use a dense vector to model their relations. It is calculated by continuous transformation of the head and tail information. Thus, we think it can not only represent the relation between two tokens, but also indicate more detailed information, such as the distance between a character and a word. Let $head[i]$ and $tail[i]$ denote the head and tail position of span $x_i$. Four kinds of relative distances can be used to indicate the relation between $x_i$ and $x_j$. They can be calculated as:
  % relative distances of $x_i$'s head and $x_j$'s head ($d^{(hh)}$), $t_i$'s head and $t_j$'s tail ($d^{(ht)}$), $t_i$'s tail and $t_j$'s head ($d^{(th)}$), $t_i$'s tail and $t_j$'s tail ($d^{(tt)}$). They are a matrix of shape [m,m] respectively, and can be calculated as:

  \vspace{-1em}
  {\small
  \begin{align}
    d^{(hh)}_{ij} &= head[i] - head[j], \\
    d^{(ht)}_{ij} &= head[i] - tail[j], \\
    d^{(th)}_{ij} &= tail[i] - head[j], \\
    d^{(tt)}_{ij} &= tail[i] - tail[j],
  \end{align}
  }%
  % where $d^{({\clubsuit\spadesuit})}_{ij}$ is a matrix of shape[m,m], m is the number of tokens. $d^{(\clubsuit\spadesuit)}_{ij}$ is the distance between $\clubsuit$ of $x_i$ and $\spadesuit$ of $x_j$ and $\clubsuit$, $\spadesuit$ denote head or tail. The final relative position encoding of spans is a simple non-linear transformation of the four distances:
  where $d^{(hh)}_{ij}$ denotes the distance between head of $x_i$ and tail of $x_j$, and other $d^{(ht)}_{ij}$, $d^{(th)}_{ij}$, $d^{(tt)}_{ij}$ have similar meanings. The final relative position encoding of spans is a simple non-linear transformation of the four distances:
  \begin{equation}\small
    % R_{ij} = \operatorname{ReLu}(W_{R} (p_{d^{(hh)}} \oplus p_{d^{(th)}_ij} \oplus p_{d^{(ht)}_{ij}} \oplus p_{d^{(tt)}_ij})),
    % \label{fusion_pos}
    R_{ij} = \operatorname{ReLU}(W_{r} (\mathbf{p}_{d^{(hh)}_{ij}} \oplus \mathbf{p}_{d^{(th)}_{ij}} \oplus \mathbf{p}_{d^{(ht)}_{ij}} \oplus \mathbf{p}_{d^{(tt)}_{ij}})),
    \label{fusion_pos}
  \end{equation}
  where $W_{r}$ is a learnable parameter, $\oplus$ denotes the concatenation operator, and $\mathbf{p}_{d}$ is calculated as in \citet{NIPS2017_7181},

\vspace{-1em}
  {\small
  \begin{align}
    \mathbf{p}_d^{(2 k)} &=\sin \left(d / 10000^{2 k / d_{model}}\right), \\
    \mathbf{p}_d^{(2 k+1)} &=\cos \left(d / 10000^{2 k / d_{model}}\right),
  \end{align}}%
  % \begin{align}
  %   p_{(d, 2 k)} &=\sin \left(d / 10000^{2 k / d_{model}}\right), \\
  %   p_{(d, 2 k+1)} &=\cos \left(j / 10000^{2 k / d}\right),
  % \end{align}}%
  where $d$ is $d^{(hh)}_{ij}$, $d^{(ht)}_{ij}$, $d^{(th)}_{ij}$ or $d^{(tt)}_{ij}$ and $k$ denotes the index of dimension of position encoding.
  % where $k$ denotes the index of dimension of position encoding.
  Then we use a variant of self-attention \cite{DBLP:journals/corr/abs-1901-02860} to leverage the relative span position encoding as follows:

\vspace{-1em}
  {\small
  \begin{align}\mathbf{A}_{i, j}^{\mathrm{*}} &={\mathbf{W}_{q}^{\top} \mathbf{E}_{x_{i}}^{\top}  \mathbf{E}_{x_{j}}} \mathbf{W}_{k, E} + \mathbf{W}_{q}^{\top} {\mathbf{E}_{x_{i}}^{\top} \mathbf{R}_{ij}} \mathbf{W}_{k, R}  \nonumber\\ &+{\mathbf{u}^{\top} \mathbf{E}_{x_{j}}} \mathbf{W}_{k, E} +{\mathbf{v}^{\top} \mathbf{R}_{ij}} \mathbf{W}_{k, R},
    \label{relative_position_eq}
  \end{align}}%
  where $\mathbf{W}_q,\mathbf{W}_{k,R},\mathbf{W}_{k,E} \in \mathbb{R}^{d_{model} \times d_{head}}$ and $\mathbf{u},\mathbf{v} \in \mathbb{R}^{d_{head}}$ are learnable parameters. Then we replace $A$ with $A^*$ in Eq.\eqref{weighted_sum_att}. The following calculation is the same with vanilla Transformer.
  % \mathbb{R}^{d_{model} \times d_{head}}
    \par After FLAT, we only take the character representation into output layer, followed by a Condiftional Random Field (CRF) \cite{Lafferty:2001:CRF:645530.655813}.

\begin{table}[]
  \small \setlength{\tabcolsep}{7.5pt}
  \begin{tabular}{@{}lcccc@{}}
  \toprule
  \multicolumn{1}{c}{} & \multicolumn{1}{l}{Ontonotes} & \multicolumn{1}{l}{MSRA} & \multicolumn{1}{l}{Resume} & \multicolumn{1}{l}{Weibo} \\ \midrule
  Train          & 15740                         & 46675                    & 3821                       & 1350                      \\
  Char$_{avg}$       & 36.92                         & 45.87                    & 32.15                      & 54.37                     \\
  Word$_{avg}$       & 17.59                         & 22.38                    & 24.99                      & 21.49                     \\
  Entity$_{avg}$          & 1.15                          & 1.58                     & 3.48                       & 1.42                      \\ \bottomrule
  \end{tabular}
  \caption{Statistics of four datasets. `Train' is the size of training set. `Char$_{avg}$', `Word$_{avg}$', `Entity$_{avg}$' are the average number of chars, words mateched by lexicon and entities in an instance. }
  \label{statistics}
  \end{table}

  \begin{table}[t]
    \centering\small \setlength{\tabcolsep}{2pt}
    \begin{tabular}{@{}l|c|cccc@{}}
      \toprule
                   & \multicolumn{1}{l|}{Lexicon} & \multicolumn{1}{l}{Ontonotes} & \multicolumn{1}{l}{MSRA} & \multicolumn{1}{l}{Resume} & \multicolumn{1}{l}{Weibo} \\ \midrule
      BiLSTM       & -                           & 71.81                         & 91.87                    & 94.41                      & 56.75                     \\
      TENER        & -                           & 72.82                         & 93.01                    & 95.25                      & 58.39                     \\
      \midrule
      Lattice LSTM & YJ                          & 73.88                         & 93.18                    & 94.46                      & 58.79                     \\
      CNNR         & YJ                          & 74.45                         & 93.71                    & 95.11                      & 59.92                     \\
      LGN          & YJ                          & 74.85                         & 93.63                    & 95.41                      & 60.15                     \\
      PLT          & YJ                          & 74.60                          & 93.26                    & 95.40                       & 59.92                     \\
      FLAT  & YJ                          & \textbf{76.45}                          & \textbf{94.12}                    & \textbf{95.45}                       & \textbf{60.32}                     \\
      FLAT$\mathrm{_{msm}}$  & YJ & 73.39 & 93.11 & 95.03 &57.98 \\
      FLAT$\mathrm{_{mld}}$  & YJ & 75.35 & 93.83 & 95.28 & 59.63 \\
      % FLAT$\mathrm{_{head}}$  & YJ                     & 75.0 &93.60 & 95.24 & 59.43 \\
      % SL           &                             & 75.54                         & 93.5                     & 95.59                      & 61.24                     \\
      \midrule
      CGN          & LS                          & 74.79                         & 93.47                    & \ \ 94.12$^*$                          & 63.09                     \\
      FLAT  & LS                          & \textbf{75.70}                          & \textbf{94.35}                    & \textbf{94.93}                          & \textbf{63.42}                      \\

      \bottomrule
      \end{tabular}
    \caption{Four datasets results (F1). BiLSTM results are from \citet{DBLP:journals/corr/abs-1805-02023}.  PLT denotes the porous lattice Transformer \cite{Mengge2019PorousLT}. `YJ' denotes the lexicon released by \citet{DBLP:journals/corr/abs-1805-02023}, and `LS' denotes the lexicon released by \citet{li-etal-2018-analogical}. The result of other models are from their original paper. %FLAT$\mathrm{_{head}}$ denotes replace $R_{ij}$ with $\mathbf{p}_{d^{(hh)}_{ij}}$  in Eq.\eqref{fusion_pos}.
    Except that the superscript * means the result is not provided in the original paper, and we get the result by running the public source code.
    Subscripts `msm' and `mld' denote FLAT with the mask of self-matched words and long distance (\textgreater 10), respectively.
    }
    \label{result-table}
    \end{table}

    % \caption{`Span' denotes using $R_{ij}$ in Eq.\eqref{fusion_pos} and `Head' denotes $p_{d^{(hh)}}$}
% \usepackage{booktabs}

\begin{table}[]
  \small
  \setlength{\tabcolsep}{4pt}
  \begin{tabular}{@{}lccllll@{}}
  \toprule
             & \multicolumn{2}{c}{Span F}                               & \multicolumn{2}{c}{Type Acc} &  &  \\ \midrule
             & \multicolumn{1}{l}{Ontonotes} & \multicolumn{1}{l}{MSRA} & Ontonotes       & MSRA       &  &  \\
  TENER      & 72.41                         & 93.17                    & 96.33           & 99.29      &  &  \\
  FLAT       & 76.23                         & 94.58                    & 97.03           & 99.52      &  &  \\
  FLAT$_{head}$ & 75.64                         & 94.33                    & 96.85           & 99.45      &  &  \\\bottomrule
  \end{tabular}
  \caption{Two metrics of models. FLAT$_{head}$ means $R_{ij}$ in \eqref{relative_position_eq} is replaced by $d^{(hh)}_{ij}$. }
  \label{fine-grained analysis}
  \end{table}

% \begin{table}[]
%   \small
%   \setlength{\tabcolsep}{7.5pt}
%   \centering
%   \begin{tabular}{@{}lll@{}}
%   \toprule
%   {Span F} & Ontonotes                 & MSRA                      \\ \midrule
%   TENER                      & \multicolumn{1}{c}{72.41} & \multicolumn{1}{c}{93.17} \\
%   FLAT                       & \multicolumn{1}{c}{76.23} & \multicolumn{1}{c}{94.58} \\
%   FLAT$_{head}$                 & \multicolumn{1}{c}{75.64} & \multicolumn{1}{c}{94.33} \\ \midrule
%                             %  &                           &                           \\ \midrule
%   Type Acc                   & Ontonotes                 & MSRA                      \\ \midrule
%   TENER                      & \multicolumn{1}{c}{96.33}                     & \multicolumn{1}{c}{99.29}                     \\
%   FLAT                       & \multicolumn{1}{c}{97.03}                     & \multicolumn{1}{c}{99.52}                     \\
%   FLAT$_{head}$                 & \multicolumn{1}{c}{96.85}                     & \multicolumn{1}{c}{99.45}                     \\ \bottomrule
%   \end{tabular}
%   \caption{Two metrics of models. FLAT$_{head}$ means $R_{ij}$ in \eqref{relative_position_eq} is replaced by $d^{(hh)}_{ij}$. }
%   \label{fine-grained analysis}
%   \end{table}
  \section{Experiments}
 \subsection{Experimental Setup}
  Four Chinese NER datasets were used to evaluate our model, including (1) \textbf{Ontonotes 4.0} \cite{trove.nla.gov.au/work/192067053} (2) \textbf{MSRA} \cite{levow-2006-third} (3) \textbf{Resume} \cite{DBLP:journals/corr/abs-1805-02023} (4) \textbf{Weibo} \cite{peng-dredze-2015-named,HeS16}.
  %where Ontonotes and MSRA are in news domain, Resume and Weibo are in social media domain.
  We show statistics of these datasets in Table \ref{statistics}. We use the same train, dev, test split as \citet{gui-etal-2019-lexicon}. We take BiLSTM-CRF and TENER \cite{Yan2019TENERAT} as baseline models. TENER is a Transformer using relative position encoding for NER, without external information. We also compare FLAT with other lexicon-based methods.
%  \subsection{Hyper-parameter settings}
   The embeddings and lexicons are the same as \citet{DBLP:journals/corr/abs-1805-02023}. When comparing with CGN~\cite{li-etal-2018-analogical}, we use the same lexicon as CGN. The way to select hyper-parameters can be found in the supplementary material. In particular, we use only one layer Transformer encoder for our model.
  % For MSRA and Ontonotes these two large dataset, we select the hyper-parameters based on the development experiment of Ontonotes. For two small datasets, Resume and Weibo, we find their optimal hyper-parameters by random-search. The hyper-parameters and the random-search range can be found in the supplementary material.
  %数据集
  %对比的模型
  %超参设计
 \subsection{Overall Performance}
  % \paragraph{Overall Performance}
  As shown in Table \ref{result-table}, our model outperforms baseline models and other lexicon-based models on four Chinese NER datasets. Our model outperforms TENER \cite{Yan2019TENERAT} by 1.72 in average F1 score. For lattice LSTM, our model has an average F1 improvement of 1.51 over it. When using another lexicon \cite{li-etal-2018-analogical}, our model also outperforms CGN by 0.73 in average F1 score. Maybe due to the characteristic of Transformer, the improvement of FLAT over other lexicon-based models on small datasets is not so significant like that on large datasets.
  %We also run our model with the same lexicon as them, and result shows our model exceeds theirs by 1.37 in average F1. %To verify the positive impact of our span position encoding, We also compare it with a common way, just encoding the first character position of tokens \cite{xiao-etal-2019-lattice}. In detail, we replace $R_{ij}$ with $p_{d^{(hh)}}$ in Eq.\eqref{fusion_pos}. The improvement of span position encoding over only using head position is obvious. To our surprise, FLAT with only head position information also outperforms TENER significantly, which may indicate our model can learn tokens' length during training.
  \subsection{Advantage of Fully-Connected Structure} %To exploit why our model outperforms lattice LSTM significantly, we use two masks on our model respectively: 1)mask attention from the character to its self-matched word 2)mask attention between two tokens, whose distance exceeds 10. Two masks correspond to two drawbacks of lattice LSTM, respectively. As shown in Table \ref{result-table}, two masks both degrade the performance of our model, and the deterioration brought by the first mask is more noticeable. As a result, we think it is important to fully encode the information of self-matched words.
  We think self-attention mechanism brings two advantages over lattice LSTM: 1) All characters can directly interact with its self-matched words. 2) Long-distance dependencies can be fully modeled. Due to our model has only one layer, we can strip them by masking corresponding attention. In detail, we mask attention from the character to its self-matched word and attention between tokens whose distance exceeds 10. As shown in Table \ref{result-table}, the first mask
\todo[inline]{多了这} brings a significant deterioration to FLAT while the second degrades performance slightly. As a result, we think leveraging information of self-matched words is important For Chinese NER.

  \begin{figure}
    {\scalefont{0.6}
    % \resizebox{0.7}{0.5}{
  \begin{tikzpicture}[]
    % \tikzstyle{every node}=[font=\footnotesize]
    \begin{axis}[
      ybar,
      scale only axis,
      yscale=0.5,
      xscale=0.75,
      enlargelimits=0.15,
      legend style={at={(0.5,-0.2)},
        anchor=north,legend columns=-1},
      ylabel={Relative Speed},
      y label style={at={(0em,10em)}},
      symbolic x coords={LatticeLSTM$^\clubsuit$,LatticeLSTM$^\spadesuit$,LR-CNN$^\clubsuit$,%
      LGN$^\spadesuit$,CGN$^\spadesuit$,FLAT$^\clubsuit$,FLAT$^\spadesuit$},
      xtick=data,
      nodes near coords,
    nodes near coords align={vertical},
      x tick label style={rotate=30,anchor=east},
      % y tick label style={},
      ]
      % ]
      \addplot coordinates {(LatticeLSTM$^\clubsuit$,1) (LatticeLSTM$^\spadesuit$,2.1)
      (LR-CNN$^\clubsuit$,2.47) (LGN$^\spadesuit$,2.74) (CGN$^\spadesuit$,4.74) (FLAT$^\clubsuit$,3.28) (FLAT$^\spadesuit$,16.32)};
    \end{axis}
  \end{tikzpicture}
    }
  \caption{Inference-speed of different models, compared with lattice LSTM $\clubsuit$. $\clubsuit$ denotes non-batch-parallel version, and $\spadesuit$ indicates the model is run in 16 batch size parallelly. For model LR-CNN, we do not get its batch-parallel version.}
  \label{speed-table}
  \vspace{-3mm}
  \end{figure}
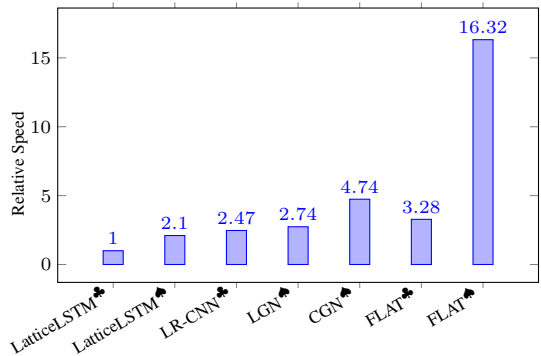

  \subsection{Efficiency of FLAT} To verify the computation efficiency of our model, we compare the inference-speed of different lexicon-based models on Ontonotes. The result is shown in Figure \ref{speed-table}. GNN-based models outperform lattice LSTM and LR-CNN. But the RNN encoder of GNN-based models also degrades their speed. Because our model has no recurrent module and can fully leverage parallel computation of GPU, it outperforms other methods in running efficiency.
  %We can see non-batch-parallelism and multiple recurrent computation make lattice LSTM be the slowest among them. Due to the complicated mechanism of padding and masking, lattice LSTM (batch\_size=16) runs slower than LR-CNN (batch\_size=1). By highly efficient graph neural network (GNN) library, GNN-based models outperform lattice LSTM and LR-CNN. LGN is beaten by CGN in speed because of its recurrent state updating. Because our model has no recurrent module and can fully leverage parallel computation of GPU, it outperforms other methods in running efficiency.
  In terms of leveraging batch-parallelism, the speedup ratio brought by batch-parallelism is 4.97 for FLAT, 2.1 for lattice LSTM, when batch size = 16. Due to the simplicity of our model, it can benefit from batch-parallelism more significantly.

  \subsection{How FLAT Brings Improvement} Compared with TENER, FLAT leverages lexicon resources and uses a new position encoding. To probe how these two factors bring improvement. We set two new metrics, 1) \textbf{Span F}: while the common F score used in NER considers correctness of both the span and the entity type, Span F only considers the former. 2) \textbf{Type Acc}: proportion of full-correct predictions to span-correct predictions. Table \ref{fine-grained analysis} shows two metrics of three models on the devlopment set of Ontonotes and MSRA. We can find: 1) FLAT outperforms TENER in two metrics significantly. 2) The improvement on Span F brought by FLAT is more significant than that on Type Acc. 3) Compared to FLAT, FLAT$_{head}$'s deterioration on Span F is more significant than that on Type Acc. These show: 1) The new position encoding helps FLAT locate entities more accurately. 2) The pre-trained word-level embedding makes FLAT more powerful in entity classification \citep{ner_analysis_1}.

  \begin{table}[t]
    \centering\small \setlength{\tabcolsep}{2pt}
    \begin{tabular}{@{}l|c|cccc@{}}
      \toprule
                   & \multicolumn{1}{l|}{Lexicon} & \multicolumn{1}{l}{Ontonotes} & \multicolumn{1}{l}{MSRA} & \multicolumn{1}{l}{Resume} & \multicolumn{1}{l}{Weibo} \\ \midrule
      BERT       & -                           & 80.14                         & 94.95                    & 95.53                      & 68.20                     \\
      BERT+FLAT        & YJ                           & 81.82                         & 96.09                    & 95.86                      & 68.55                     \\
      \bottomrule
      \end{tabular}
      \caption{Comparision between BERT and BERT+FLAT. `BERT' refers to the BERT+MLP+CRF architecture. `FLAT+BERT' refers to FLAT using BERT embedding. We finetune BERT in both models during training. The BERT in the experiment is `BERT-wwm' released by \citet{DBLP:journals/corr/abs-1906-08101}. We use it by the BERTEmbedding in fastNLP \protect\footnotemark .}
      \label{BERT-result}

    \end{table}
    \footnotetext{https://github.com/fastnlp/fastNLP}

  \subsection{Compatibility with BERT} We also compare FLAT equipped with BERT with common BERT+CRF tagger on four datasets, and Results are shown in Table \ref{BERT-result}. We find that, for large datasets like  Ontonotes and MSRA, FLAT+BERT can have a significant improvement over BERT. But for small datasets like Resume and Weibo, the improvement of FLAT+BERT over BERT is marginal. %Maybe for small datasets, injecting word information into BERT \cite{yubowen} is more proper.

  \section{Related Work}
  % \subsection{Chinese NER with Lexicon}
  % Some comparison has been done between character-based and word-based model for Chinese NER \cite{li-etal-2014-comparison}, And reflects that with the limited performance of the current Chinese word segmentation, word-based models can be easily beaten by character-based counterparts.
  % In this section, we summarize the related work.
  \subsection{Lexicon-based NER} \citet{DBLP:journals/corr/abs-1805-02023} introduced a lattice LSTM to encode all characters and potential words recognized by a lexicon in a sentence, avoiding the error propagation of segmentation while leveraging the word information. \citet{Gui:2019:CCN:3367722.3367738} exploited a combination of CNN and rethinking mechanism to encode character sequence and potential words at different window sizes. Both models above suffer from the low inference efficiency and are hard to model long-distance dependencies. \citet{gui-etal-2019-lexicon} and \citet{sui-etal-2019-leverage} leveraged a lexicon and character sequence to construct graph, converting NER into a node classification task. However, due to NER's strong alignment of label and input, their model needs an RNN module for encoding. The main difference between our model and models above is that they modify the model structure according to the lattice, while we use a well-designed position encoding to indicate the lattice structure.
  % Recently, \citet{Mengge2019PorousLT} proposed a variant of transformer to encode lattice input, which also needs an RNN to decode.
  \todo[inline]{Different}

  % \subsection{Lattice-Based Transformer}
  % 感觉self-attention的讲解，如果在篇幅不够的情况下可以删除
  % As a self-attention model, transformer is used for modeling sequence and has an ability of global encoding. It realizes decoupling the representation of token and position, which enables its high computation efficiency. Different from vanilla transformer, lattice-based transformer has a lattice-structure input. And how to adjust position encoding is a main difficulty because lattice is structured like graph.
  \subsection{Lattice-based Transformer} For lattice-based Transformer, it has been used in speech translation and Chinese-source translation. The main difference between them is the way to indicate lattice structure. In Chinese-source translation, \citet{xiao-etal-2019-lattice} take the absolute position of nodes' first characters and the relation between each pair of nodes as the structure information. In speech translation, \citet{sperber-etal-2019-self} used the longest distance to the start node to indicate lattice structure, and \citet{zhang-etal-2019-lattice} used the shortest distance between two nodes. Our span position encoding is more natural, and can be mapped to all the three ways, but not vise versa. Because NER is more sensitive to position information than translation, our model is more suitable for NER. Recently, Porous Lattice Transformer \cite{Mengge2019PorousLT} is proposed for Chinese NER. The main difference between FLAT and Porus Lattice Transformer is the way of representing position information. We use `head' and `tail' to represent the token's position in the lattice. They use `head', tokens' relative relation (not distance) and an extra GRU. They also use `porous' technique to limit the attention distribution. In their model, the position information is not recoverable because `head' and relative relation can cause position information loss. Briefly, relative distance carries more information than relative relation.
  % But these models are designed for seq2seq task, which is sentence-alignment, while NER is token-alignment task. They are not suitable for NER.
  \todo[inline]{Different}

  \section{Conclusion and Future Work}
  In this paper, we introduce a flat-lattice Transformer to incorporate lexicon information for Chinese NER. The core of our model is converting lattice structure into a set of spans and introducing the specific position encoding. Experimental results show our model outperforms other lexicon-based models in the performance and efficiency. We leave adjusting our model to different kinds of lattice or graph as our future work.

  \section*{Acknowledgments}
  We thank anonymous reviewers for their responsible attitude and helpful comments. We thank Tianxiang Sun, Yunfan Shao and Lei Li for their help, such as drawing skill sharing, pre-reviewing, etc. This work is supported by the National Key Research and Development Program of China (No. 2018YFC0831103), National Natural Science Foundation of China (No. U1936214 and 61672162), Shanghai Municipal Science and Technology Major Project (No. 2018SHZDZX01) and ZJLab.

  \bibliography{acl2020}
  \bibliographystyle{acl_natbib}

  \section{Appendices}
  \label{sec:appendix}
  \subsection{Hyperparameters Selection}
For MSRA and Ontonotes these two large datasets, we select the hyper-parameters based on the development experiment of Ontonotes. For two small datasets, Resume and Weibo, we find their optimal hyper-parameters by random-search. The Table \ref{hyper-big} lists the hyper-parameters obtained from the development experiment of Ontonotes.
\begin{table}[h]
  \centering
  \begin{tabular}{@{}ll@{}}
  \toprule
  batch                                                           & 10                                                     \\ \midrule
  \begin{tabular}[c]{@{}l@{}}lr\\   \ -decay\end{tabular}           & \begin{tabular}[c]{@{}l@{}}1e-3\\   0.05\end{tabular} \\
  \begin{tabular}[c]{@{}l@{}}optimizer\\   \ -momentum\end{tabular} & \begin{tabular}[c]{@{}l@{}}SGD\\   0.9\end{tabular}    \\
  d$_model$                                                        & 160                                                    \\
  head                                                            & 8                                                      \\
  FFN size                                                        & 480                                                    \\
  embed dropout                                                   & 0.5                                                    \\
  output dropout                                                  & 0.3                                                    \\
  warmup                                                          & 10 (epoch)                                               \\ \bottomrule
  \end{tabular}
  \caption{Hyper-parameters for Ontonotes and MSRA.}
  \label{hyper-big}
  \end{table}

% \usepackage{booktabs}
The Table \ref{random-search} lists the range of hyper-parameters random-search for Weibo, Resume datasets. For the hyper-parameters which do not appear in it, they are the same as in Table \ref{hyper-big}.
\begin{table}[h]
  \centering
  \begin{tabular}{@{}ll@{}}
  \toprule
  batch   & {[}8,10{]}             \\ \midrule
  lr      & {[}1e-3, 8e-4{]}       \\
  d$_head$ & {[}16,20{]}            \\
  head    & {[}4,8,12{]}           \\
  warmup  & {[}1, 5, 10{]} (epoch) \\ \bottomrule
  \end{tabular}
  \caption{The range of hyper-parameters random-search for Weibo, Resume datasets.}
  \label{random-search}
  \end{table}

  % \section{Appendix}
  % \subsection{Hyper-parameters}
  % % Please add the following required packages to your document preamble:
  % % \usepackage{booktabs}
  % \begin{table}[]
  %   \centering
  %   \begin{tabular}{@{}llll@{}}
  %   \toprule
  %   parameter       & value & parameter     & value   \\ \midrule
  %   batch           & 10    & lr            & 1e-3    \\
  %   optimizer       & SGD   & momentum      & 0.9     \\
  %   char emb size   & 50    & word emb size & 50      \\
  %   bigram emb size & 50    & warmup        & 1 epoch \\
  %   d\_model        & 128   & head          & 8       \\
  %   FFN size        & 384   &               &         \\ \bottomrule
  %   \end{tabular}
  %   \end{table}
  \end{CJK}

\end{document}